\documentclass[conference]{IEEEtran}
\IEEEoverridecommandlockouts
\usepackage{cite}
\usepackage{amsmath,amssymb,amsfonts}
\usepackage{algorithmic}
\usepackage{graphicx}
\usepackage{textcomp}
\usepackage{xcolor}
\usepackage{colortbl}
\usepackage{balance}
\usepackage{flushend}
\usepackage{multirow}
\usepackage[utf8]{inputenc}
\usepackage[hyphens]{url}
\usepackage{hyperref}
\usepackage{hyperref}
\usepackage{subfigure}
\usepackage{comment}

\usepackage{booktabs}
\usepackage{tabularx}
\newcolumntype{Y}{>{\centering\arraybackslash}X}

\usepackage{fancyhdr}
\DeclareGraphicsExtensions{.pdf,.svg,.jpg,.png,.eps} % first try pdf, then eps, png and jpg
\graphicspath{{./assets/}} % Path to a folder where all pictures are located

\def\BibTeX{{\rm B\kern-.05em{\sc i\kern-.025em b}\kern-.08em
    T\kern-.1667em\lower.7ex\hbox{E}\kern-.125emX}}

\begin{document}

\title{KutralNet: A Portable Deep Learning\\Model for Fire Recognition}

\author{\IEEEauthorblockN{
Angel Ayala\,$^{1}$, 
Bruno Fernandes\,$^{1}$, 
Francisco Cruz\,$^{2,3}$, \\
David Macêdo\,$^{4,5}$, 
Adriano L. I. Oliveira\,$^{4}$, and
Cleber Zanchettin\,$^{4,6}$
}
\IEEEauthorblockA{\,
$^{1}$Escola Polit\'ecnica de Pernambuco, Universidade de Pernambuco, Recife, Brasil\\
$^{2}$School of Information Technology, Deakin University, Geelong, Australia\\
$^{3}$Escuela de Ingenier\'ia, Universidad Central de Chile, Santiago, Chile\\
$^{4}$Centro de Inform\'atica, Universidade Federal de Pernambuco, Recife, Brasil\\
$^{5}$Montreal Institute for Learning Algorithms, University of Montreal, Quebec, Canada\\
$^{6}$Department of Chemical and Biological Engineering, Northwestern University, Evanston, United States of America\\
Emails: \{aaam, bjtf\}@ecomp.poli.br, francisco.cruz@deakin.edu.au, \{dlm, alio, cz\}@cin.ufpe.br}
}

\maketitle

\begin{abstract}
Most of the automatic fire alarm systems detect the fire presence through sensors like thermal, smoke, or flame. 
One of the new approaches to the problem is the use of images to perform the detection. 
The image approach is promising since it does not need specific sensors and can be easily embedded in different devices. 
However, besides the high performance, the computational cost of the used deep learning methods is a challenge to their deployment in portable devices.
In this work, we propose a new deep learning architecture that requires fewer floating-point operations (flops) for fire recognition.
Additionally, we propose a portable approach for fire recognition and the use of modern techniques such as inverted residual block, convolutions like depth-wise, and octave, to reduce the model's computational cost.
The experiments show that our model keeps high accuracy while substantially reducing the number of parameters and flops.
One of our models presents 71\% fewer parameters than FireNet, while still presenting competitive accuracy and AUROC performance.
The proposed methods are evaluated on FireNet and FiSmo datasets. 
The obtained results are promising for the implementation of the model in a mobile device, considering the reduced number of flops and parameters acquired.
\end{abstract}

\thispagestyle{fancy}
%%%%%%%%%%%%%%%%%%%%%%%%%%%%%%%%%%%%%%%%%%%%%%%%%%%%%%%%%%%%%%%%%%%%%%%%%%%%%%%%%%%%%%%%%%%%%%%%%%%%%%%%%%%%%%%%%%%%%%%%%%%%%%%%%%%%%%%%%%%%%%%%%%%%
\section{Introduction}

The fire alarm systems are a combination of sensors and machine learning algorithms to identify patterns of warning. 
Usually, the sensor system is composed of: (i) heat or temperature detectors, which typically are not early warning devices; (ii) flame detectors that habitually are built-on optical, UV, and IR sensors; and (iii) smoke detectors, frequently using photoelectric, ionization, or a combination of both. 
The use of images to fire recognition is a new promising approach \cite{Alves2019:FireImage}, based on the excellent results that deep learning models are obtaining in image processing applications \cite{Chollet2017:Xception} allowing to avoid the use of special sensors to perform the recognition. 

The Deep Learning (DL) approach \cite{LeCun2015:deeplearning} has proved to be suitable for automating the feature acquisition from complex data in machine learning tasks. 
In such a way, DL works in multiple levels of abstraction for data representation. 
Considering that, the use of computer vision to fire recognition reduces the necessity of specific sensors, being suitable for the inlay to portable, remote, and mobile devices.
However, DL approaches have some challenges: (i) the required computational resources; (ii) the model's computation complexity and size; and (iii) the quantity of data needed for its training, among others.
The previously mentioned challenges can be tackled by focusing on the development of deep learning models for mobile devices \cite{Deng2019:DLMobile}, which has been a less explored area in DL literature. 
For robotics or autonomous systems being portables, they must be capable of working with limited processing power, storage, and energy, besides providing an efficient inference time.

\begin{figure*}%[!t]
    \centering
    % Inverted residual block
    \subfigure[The inverted residual block.
    Diagonally hatched layers do not use non-linearities.
    The thickness of each block is used to indicate its relative number of channels. 
    The inverted residuals connect the bottlenecks.
    Adapted from \cite{Sandler2018:MobileNet}.
    ]{\includegraphics[width=0.99\columnwidth]{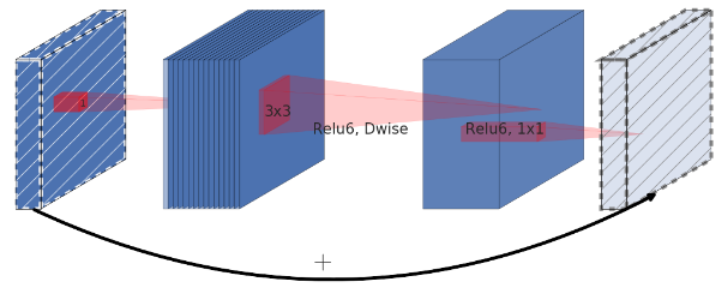}
    \label{fig:InvertedResidualBlock}}
    \hspace{0.1cm}
    \subfigure[Detailed design of the octave convolution.
        Green arrows correspond to information updates, while red arrows facilitate information exchange between the two frequencies.
        Adapted from \cite{Chen2019:octave}.
        ]{\includegraphics[width=0.99\columnwidth]{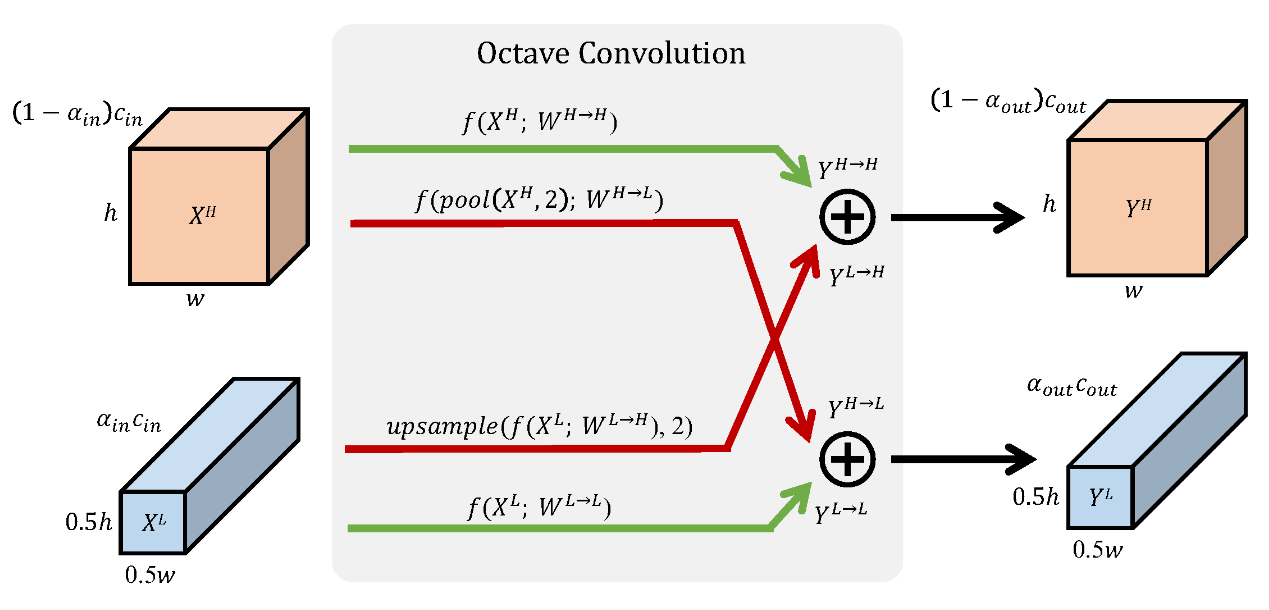}
        \label{fig:OctaveConvolution}
        }
    \caption{The fundamental convolutional blocks used in this work.}
    \label{fig:AllBlocks}
\end{figure*}

Some works have addressed portable models of deep learning using more efficient ways to compute the convolutions, such as presented by Sandler et al. \cite{Sandler2018:MobileNet}. 
The authors proposed the inverted residual block (see Figure \ref{fig:InvertedResidualBlock}) with point-wise and depth-wise convolutions to simplify the dimensionality of the signal processing. 
Thus, the proposed deep model offers an interesting trade-off between accuracy, the number of operations, and the number of parameters considering other models such as ShuffleNet \cite{Zhang2018:ShuffleNet} and NasNet \cite{Zoph2018:NasNet}.
Another efficient approach for portable models is the work presented by Chen et al. \cite{Chen2019:octave}, the octave convolution (see Figure \ref{fig:OctaveConvolution}), which reduces the spatial redundancy of the signal, separating into a high and low spatial frequency for processing. 
In this case, the size of the separation depends on an $\alpha$ ratio, which defines the \textit{octave feature representation} factorizing the feature map into groups for each frequency. 
The $\alpha$ value also establishes the quantity of memory, operations, and the number of parameters reduced for the model.

% COMBINED Inverted residual block + Octave

In this paper, we propose a new deep learning model for fire recognition called KutralNet\footnote{We took inspiration from Mapuche language or Mapudungun where k\"utral means fire.}, which comprises five layers and require 92\% fewer floating-point operations (flops) for processing in comparison with previous approaches.
This model is used as a baseline to build portable models that compare the efficiency in signal processing of the inverted residual block, the depth-wise convolution, and octave convolution approaches.
Our best resultant model, KutralNet Mobile Octave, presents a competitive validation accuracy and AUROC performance despite using 71\% fewer numbers of parameters in comparison with FireNet.
We compare the proposed models with state-of-the-art approaches to fire recognition over the FireNet and FiSmo datasets.

In Section II, we present related works. Section III the proposed model. Section IV presents the experiment setup and in Section V, the results and discussion. Section VI presents the final remarks.

%%%%%%%%%%%%%%%%%%%%%%%%%%%%%%%%%%%%%%%%%%%%%%%%%%%%%%%%%%%%%%%%%%%%%%%%%%%%%%%%%%%%%%%%%%%%%%%%%%%%%%%%%%%%%%%%%%%%%%%%%%%%%%%%%%%%%%%%%%%%%%%%%%%%
\section{Related Works}
In the last years, multiple methods to automate fire recognition were proposed, most of them from video surveillance systems, such as Closed Circuit TV systems. 
Some surveillance equipment uses cameras with low resolution, at a flat frame rate, or with no storage option, and others more sophisticated ones with proper image resolution cameras.

The first approaches to fire recognition in computer vision were addressed using techniques based on RGB color space \cite{YoonHo2014:rgbfire}, spectral color \cite{Nikos2011:sensor}, texture recognition \cite{Dimitropoulos2015:flame}, and spatio-temporal treatment \cite{Barmpoutis2013:video}.
The most recent methods correspond to deep neural networks, especially convolutional neural networks (CNNs). 
Many of the DL implementations \cite{Sharma2017:deepfire, Muhammad2018:deepfire, Namozov2018:fire} were built on previously trained models such as ResNet and its variations \cite{Kaiming2015:ResNet}.
Recently, it has been widespread the use of the DL method combined with fine-tuning or transfer-learning techniques.
This complement solves the lack of data for training deep models; the only issue is the size and complexity constraint of the network.

A new lightweight model, FireNet, was proposed by Jadon et al. \cite{Jadon2019:FireNet}, who present a dataset for training and test a model from scratch considering a fire and smoke detection system.
The model's architecture presents three consecutive convolution blocks before the classifier, and each convolution block contains a convolution layer, an average pooling layer, and a dropout normalization layer.
The classifier is composed of three fully connected hidden layers and is capable of processing images with a maximum size of 64x64 pixels on RGB channels.
Other model trained from scratch is OctFiResNet \cite{Ayala2019:OctFiResNet}.
This model is based on the initial blocks of ResNet, replacing the vanilla convolution with the octave convolution.
For this model, the optimizer corresponds to the Adam algorithm with Nesterov momentum and hyper-parameters settled to $\alpha=10^{-4}$ and $\epsilon=10^{-7}$, to process images of 96x96 pixels.
Likewise, a well-known deep learning model, ResNet \cite{Kaiming2015:ResNet} with its depth-dependent variations, is used in this task.
In this case, the ResNet50's architecture presents a slight modification of the classifier on the top of the model \cite{Sharma2017:deepfire}, using transfer-learning over a pre-trained model where the first layers perform the feature mapping of an image of 224x224 pixels.

Currently, most of the models developed using DL techniques are focused on the results, leaving in second place the resources needed for the execution of the algorithm.
The development of algorithms for mobile devices must be focused on the duration of the battery, optimizing its performance and autonomy.
A DL model must be efficient enough to work in real-time to be suitable for running in a hardware-restricted system or mobile devices.
% Madhevan does not propose the model, propose a wireless mobile surveillance robot 
For example, the development of the fire recognition DL model can work in the mobile vehicle system for fire detection proposed by Madhevan et al. \cite{Madhevan2017:mobile}.

%%%%%%%%%%%%%%%%%%%%%%%%%%%%%%%%%%%%%%%%%%%%%%%%%%%%%%%%%%%%%%%%%%%%%%%%%%%%%%%%%%%%%%%%%%%%%%%%%%%%%%%%%%%%%%%%%%%%%%%%%%%%%%%%%%%%%%%%%%%%%%%%%%%%
\section{The KutralNet Architecture}
The KutralNet model proposal for fire recognition is intended to reduce the complexity of a deep learning model to process an image and decides if it has or not fire presence.
This proposal sets a baseline model in order to develop portable models that are well suited for limited hardware devices.

% KutralNet arch
\begin{figure}%[!t]
    \centering
        \includegraphics[width=\columnwidth]{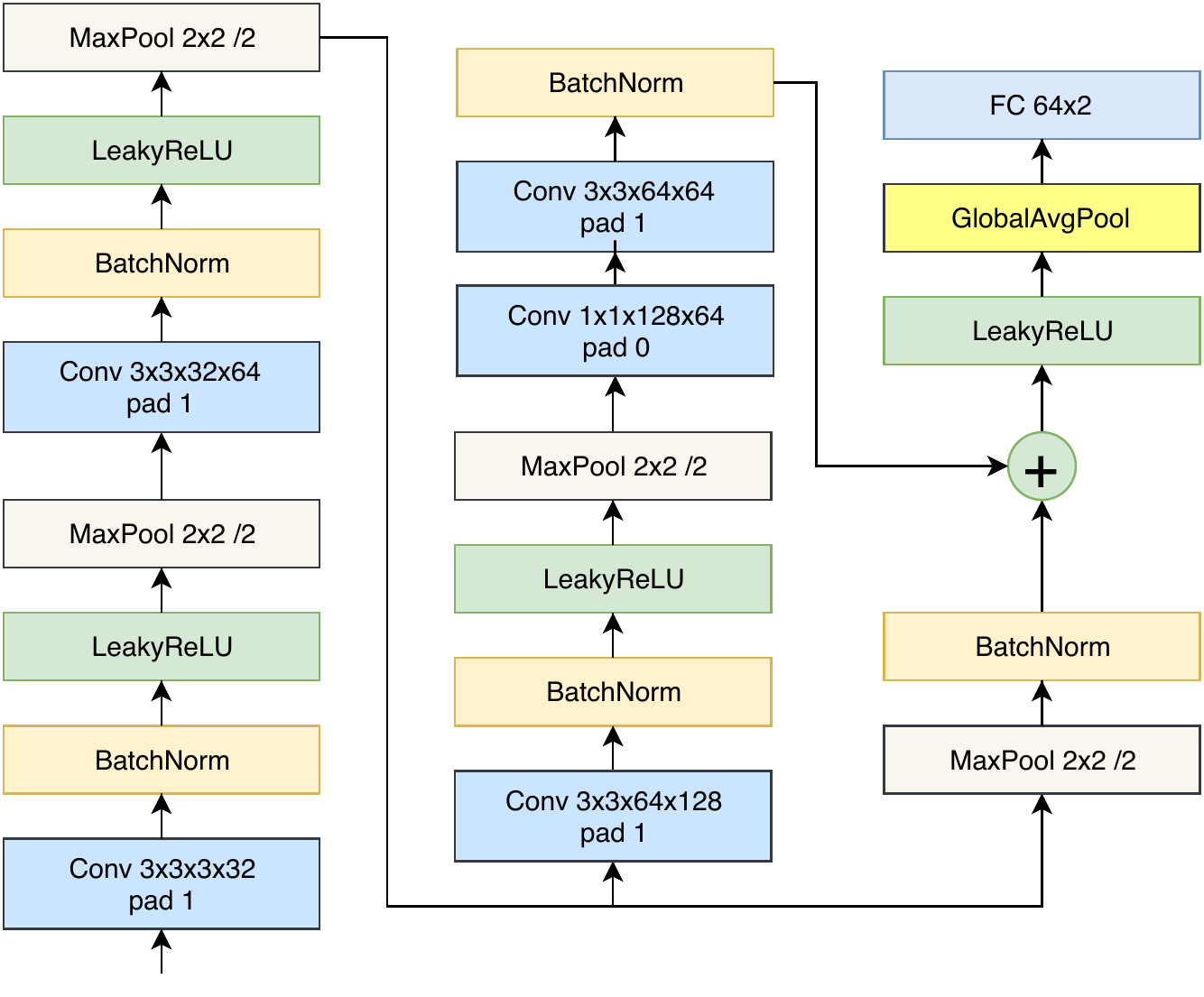}
        \caption{The KutralNet architecture works with images of 84x84 pixels in RGB channels.
        Only the first three convolution layer blocks keep the image dimensions.
        To the classifier, a global average pooling delivers the features to the fully connected (FC) layer with two outputs, one for the fire label and another for the no-fire label.
        Consecutively, a softmax function is implemented as activation at the top of the network.}
    \label{fig:KutralNet}
\end{figure}

%%%%%%%%%%%%%%%%%%%%%%%%%%%%%%%%%%%%%%%%%%%%%%%%%%%%%%%%%%%%%%%%%%%%%%%%%%
\subsection{Baseline model's architecture}
The baseline for the KutralNet model is inspired by FireNet, OctFiResNet, and the modified ResNet50 models.
Our KutralNet is the result of mixing between a deep model and a lightweight one, capable of processing 84x84 pixels images in RGB channels.
The model can be seen in Figure \ref{fig:KutralNet}.
The first three blocks consist of convolution with no bias and a 3x3 filters layer, followed by a batch-normalization layer, continuing with a LeakyReLU activation, and finally, a max-pooling with kernel 2x2 and stride 2.
When the signal passes from one block to another, it increases the number of the filters and reduces the dimension.
For the last block, two convolution layers and a batch-normalization layer are present.
The first convolution of 1x1 reduces the number of filters, and the second convolution of 3x3 processes the filters finishing with 64 channels.
A shortcut, of a 2x2 max-pooling layer with stride 2 and a batch-normalization layer, connects from the second block with the final convolution block.
On the top of the layer after the shortcut, the signal passes through a LeakyReLU activation and a global average pooling layer to the classifier, which consists of a fully connected layer with two neurons in the exit.
This architecture is defined for processing low-dimension images in a lightweight configuration.
It has been proved that few layers are capable of acquiring enough features for fire classification in order to optimize the inference time \cite{Jadon2019:FireNet}.
Additionally, using shortcut and batch-normalization layers avoids overfitting the model \cite{Kaiming2015:ResNet}.
We have chosen LeakyReLU since a non-zero slope for negative part improves the results \cite{Xu2015:ReLUActivations} and presents a low-cost implementation.

% KutralNet Mobile block
\begin{figure*}%[ht]
    \centering
    \includegraphics[width=0.8\textwidth]{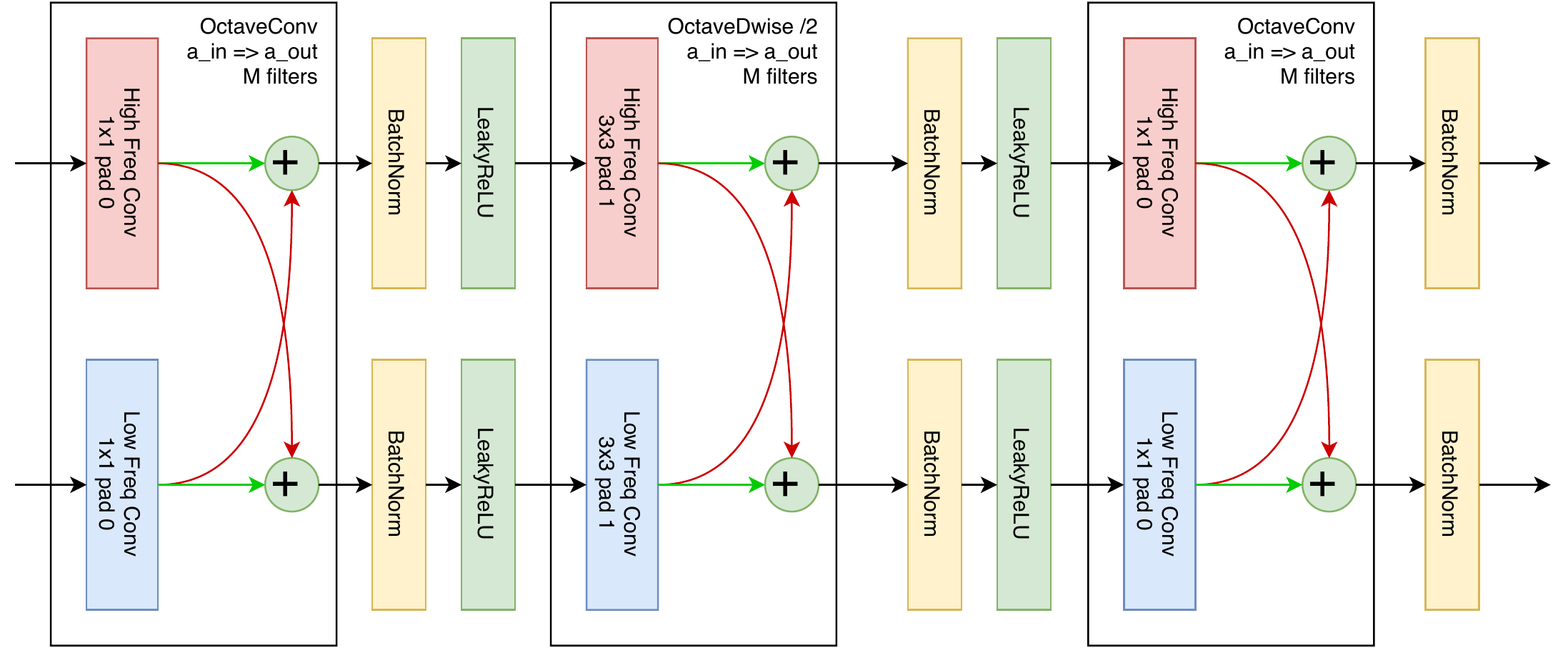}
    \caption{KutralNet Mobile Octave model resultant block.
    The most to left and right of the block present a point-wise convolution and, in the middle, the depth-wise convolution, all combined with the octave convolution with $\alpha=0.5$.}
    \label{fig:KutralNet_mobile_block}
\end{figure*}

The models' summary with the parameter numbers and operations required for image processing is in Table \ref{table:Comparative}.
The flops value for the FireNet model is not presented due to the instability during measurement, which increase the value over each run.

%Tabla comparativa baseline 
\begin{table}%[!b]
    \caption{The computational cost of each implemented model represented as parameters and flops.}
    \centering
    \begin{tabularx}{\columnwidth}{lYY}
        \toprule
        \multicolumn{1}{c}{\textbf{Model$_{(InputSize)}$}} & \multicolumn{1}{c}{\textbf{Parameters}} & \multicolumn{1}{c}{\textbf{Flops}} \\
        \midrule
        ResNet50$_{(224x224)}$ & 31.91M & 4.13G \\ 
        OctFiResNet$_{(96x96)}$ & 956.23K & 928.95M \\
        FireNet$_{(64x64)}$ & 646.82K & - \\ 
        KutralNet$_{(84x84)}$ & 138.91K & 76.85M \\
        \bottomrule
    \end{tabularx}
    \label{table:Comparative}
\end{table}

%%%%%%%%%%%%%%%%%%%%%%%%%%%%%%%%%%%%%%%%%%%%%%%%%%%%%%%%%%%%%%%%%%%%%%%%%%
\subsection{Portable version implementations}
The KutralNet model is the baseline used here to develop portable models, focusing on reducing the model size and computational cost.
The octave and depth-wise convolution \cite{Chen2019:octave,Sandler2018:MobileNet} demonstrate excellent performance with a sharp reduction of operations and parameters required, resulting in more efficient models.
This reduction is resulting from convolutions with low kernel dimensions for both cases.
For the depth-wise convolution type, the filter channels are processed in groups where $groups = C_{in}$ and $out\_channels = C_{in} * K$, in which the output filters are K times the input filters, reducing in this way the mathematical complexity of the operation.
For the octave convolution, the separate ways to process the filters on high and low frequency computing the parameters information $W$ into two components $W=[W_H, W_L]$ and exchanging information between them.
Additionally, these convolution techniques, used in different deep learning model architectures, and various tasks such as classification, object detection, and semantic segmentation, achieve a model size reduction, less computational requirements, and a slightly better performance in some cases.
This is useful for our purpose, and we present a new type of convolution combining the depth-wise convolution with its group filter operations and the octave convolution, which achieves a valuable trade-off between accuracy, model size, and computational cost.
Our different portable versions implemented are as follows:

\begin{itemize}
    \item \textbf{KutralNet Mobile:} It is inspired by MobileNetV2 \cite{Sandler2018:MobileNet} and presents the implementation of the inverted residual block.
    In this approach, from the second block, the KutralNet convolution blocks were replaced with the inverted residual block, in which each block contains point-wise and depth-wise convolution with shortcut connections in some cases.
    \item \textbf{KutralNet Octave:} It is based on the KutralNet's architecture, and all the vanilla convolution were replaced with octave convolution with an $\alpha$ parameter of $0.5$.
    Thus, the octave convolution uses the 50\% for the \textit{octave feature representation}, which corresponds to the low-frequency channel dealing with global features, and the rest for the high-frequency channel dealing with specific features.
    Additionally, the octave convolution works using the depth-wise convolution form where it is possible.
    \item \textbf{KutralNet Mobile Octave:} It is the combination of the MobileNetV2 block and the octave convolution.
    It is the same KutralNet Mobile but replacing the vanilla convolution with the octave convolution combined with depth-wise convolution form.
    The resultant block can be seen in Figure \ref{fig:KutralNet_mobile_block}.
\end{itemize}

All the portable models present the same classifier on the top of the network, which is composed of a LeakyReLU activation, passing through a global average pooling layer directly to a fully connected layer with two neurons on the exit.
Additional details of the implementations can be seen on the project repository\footnote{Github repository \url{https://github.com/angel-ayala/kutralnet}}.
A summary of the parameters and operations of the KutralNet models can be seen in Table \ref{table:PortableComparative}.
The number of floating-point operations (flops) and the number of parameters for a model are the metrics defined to measure the model requirements for image processing and storage, respectively.
The fewer parameters, the less on-disk size, is required.
Moreover, as fewer the number of flops, less is the computational cost for processing.
As the focus of our work is on develop a mobile deep learning model, the less value on both metrics, the best suited is the model for this purpose.
For the case of the Mobile and Mobile Octave variations, the models present a higher number of parameters but with a high flops reduction than the baseline.

%Tabla comparativa modelos portables
\begin{table}%[!b]
    \caption{The computational cost of each KutralNet portable variation represented as parameters and flops.}
    \centering
    \begin{tabularx}{\columnwidth}{lYY}
        \toprule
        \multicolumn{1}{c}{\textbf{Model$_{(InputSize)}$}} & \multicolumn{1}{c}{\textbf{Parameters}} & \multicolumn{1}{c}{\textbf{Flops}} \\
        \midrule
        KutralNet$_{(84x84)}$ & 138.91K & 76.85M \\ 
        KutralNet Mobile$_{(84x84)}$ & 173.09K & 43.27M \\ 
        KutralNet Octave$_{(84x84)}$ & 125.73K & 29.98M \\ 
        KutralNet Mobile Octave$_{(84x84)}$ & 185.25K & 24.59M \\
        \bottomrule
    \end{tabularx}
    \label{table:PortableComparative}
\end{table}

%%%%%%%%%%%%%%%%%%%%%%%%%%%%%%%%%%%%%%%%%%%%%%%%%%%%%%%%%%%%%%%%%%%%%%%%%%%%%%%%%%%%%%%%%%%%%%%%%%%%%%%%%%%%%%%%%%%%%%%%%%%%%%%%%%%%%%%%%%%%%%%%%%%%
\section{Experimental Setup}
The environment used for the experiments was Google Colab, an online open cloud platform for machine learning algorithms.
This online platform provides a ready to use ecosystem with all the required libraries installed, e.g., for data manipulation, data visualization, and for the training process.
The container of the environment was a virtual machine configured with up to 13GB of memory, an Intel Xeon@2.30GHz, and an Nvidia Tesla K80 with 12GB of GPU memory.

Our first experiment aimed to define a baseline model and prove its effectiveness to fire recognition.
For this purpose, three different models were implemented for comparing the baseline.
The first model is a novel lightweight model for fire recognition, the FireNet model \cite{Jadon2019:FireNet}, which is implemented in a Raspberry Pi as part of a fire alarm system.
A second model is a modified version of ResNet50 \cite{Sharma2017:deepfire} used with the transfer-learning technique for training the classifier on the top of the network with fire and no-fire images.
Another implemented model is OctFiResNet \cite{Ayala2019:OctFiResNet}, a proposal to lightweight and efficient model, based on the ResNet model, with few layers and octave convolutions.
The final implemented model is our proposal, KutralNet, to address a computationally efficient and lightweight deep neural network, balancing between the parameters and effectiveness. 
Just for the case of KutralNet, during the training stage, present a learning rate variation from $\alpha=10^{-4}$ to $\alpha=10^{-5}$ on the epoch 85.
The trained and evaluated KutralNet is the baseline for the portable approaches.

%mobile
With the defined baseline, the next set of experiments aimed to reduce the operations required for processing the images for fire recognition of the KutralNet model.
In order to reduce the operations required, techniques such as the depth-wise convolution presented with an inverted residual block in \cite{Sandler2018:MobileNet} and the octave convolution presented by Chen et al. \cite{Chen2019:octave} were implemented separately at first and mixed later to check their compatibility and efficiency.
This reduction results in three different models named: (i) KutralNet Mobile for the one where uses the inverted residual block, inspired in MobileNetV2;
(ii) KutralNet Octave is called the model which uses the octave convolution with a reduction parameter $\alpha=0.5$ combined with the depth-wise convolution, requiring fewer operations and space to store the model;
and (iii) KutralNet Mobile Octave, which uses the inverted residual block in combination with the octave convolution using the depth-wise form.

The training of all the models was performed during 100 epochs to choose the model with the best validation accuracy.
All of them were trained using cross-entropy loss and the Adam optimizer with default parameters, except for those previously mentioned.
For testing, two metrics were used to compare the efficiency of each model, the Receiver Operating Characteristics (ROC) curve and the area under the ROC curve (AUROC).

%%%%%%%%%%%%%%%%%%%%%%%%%%%%%%%%%%%%%%%%%%%%%%%%%%%%%%%%%%%%%%%%%%%%%%%%%%
\subsection{Datasets}
% dataset
For the training, validation, and test of the models, two datasets were used in this work.
The first one is called FireNet as the model \cite{Jadon2019:FireNet} and contains training and test subsets, with 2425 and 871 images, respectively.
The second one is the FiSmo dataset proposed by Cazzolato et al. \cite{Cazzolato2017:fismo}, which has been recently published with a total of 6063 images.
Additionally, we have used a contained subset of FiSmo comprised of 1968 images equally balanced between the fire and no-fire label.
An augmented version of FiSmo is also used, adding 485 black images labeled as no-fire, in order to check out the models' response to this kind of augmentation.
In addition to the balanced FiSmo version, we have also used an augmented version of this subset, which replaces 98 no-fire images for black images.
More details of the dataset images are in Table \ref{table:DatasetsSummary}, where FiSmoA is the augmented version of FiSmo, FiSmoB is the balanced version of FiSmo, and FiSmoBA is the augmented balanced version of FiSmo.

%Tabla con separación de dataset
\begin{table}%[!t]
    \caption{Quantity of images per label present on the particular dataset presented for this work.}
    \centering
    \begin{tabularx}{\columnwidth}{lYYY}
        \toprule
        \textbf{Dataset} & \textbf{Fire} & \textbf{No Fire} & \textbf{Total} \\
        \midrule
        \textbf{FireNet (training)} & 1124 & 1301 & 2425 \\ 
        \textbf{FireNet (testing)} & 593 & 278 & 871 \\ 
        \textbf{FiSmo} & 2004 & 4059 & 6063 \\ 
        \textbf{FiSmoA} & 2004 & 4544 & 6548 \\ 
        \textbf{FiSmoB} & 984 & 984 & 1968 \\ 
        \textbf{FiSmoBA} & 984 & 984 & 1968 \\ 
        \bottomrule
    \end{tabularx}
    \label{table:DatasetsSummary}    
\end{table}

The datasets were used with the same specifications presented in their original works.
For the case of FireNet, the training dataset is split into 70\% for training and 30\% for validation, with a whole new dataset included for test purposes.
For FiSmo dataset, an arbitrary split value of 80\% for training and 20\% for validation separates the images of the dataset, due to only a single implementation for model training has been found.
The augmented version presents the same dataset separation, from the 485 black images inserted, 388 correspond to training and 97 for validation.
For preprocessing, the images were rescaled to the image input size of each model and normalized with values $\in [0,1]$.

% Baseline training Results
\begin{figure*}%[htp]
    \centering
    \subfigure[The validation accuracy obtained from each dataset by the different models. The ResNet50 version works better with the augmented version of FiSmo and for the FireNet dataset, followed by KutralNet.
    Additionally, Kutralnet performs better with FiSmo.
    ]{\includegraphics[width=0.8\columnwidth]{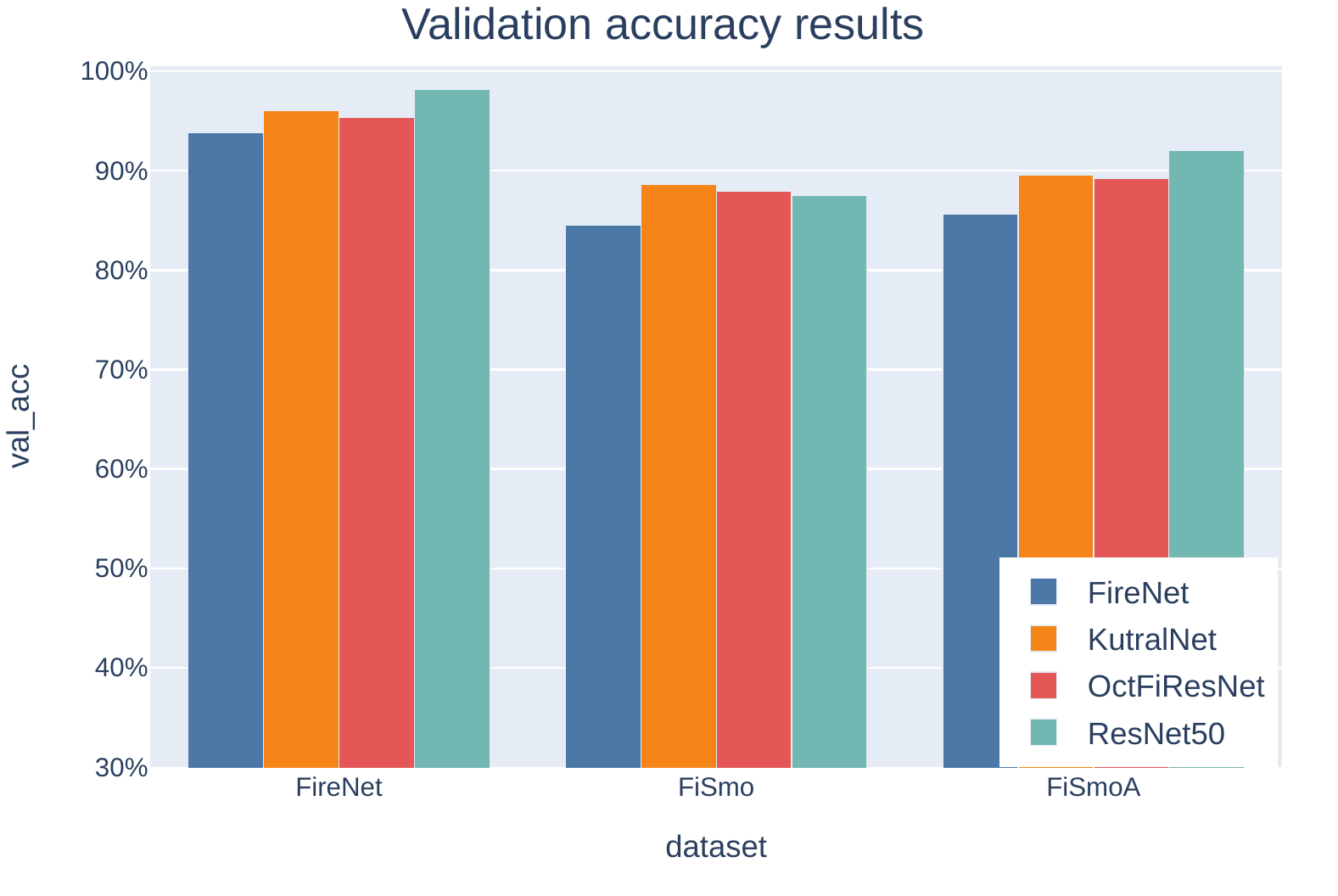}
    \label{fig:TrainingResults}
    }\quad
    \subfigure[Test accuracy of each model trained with a different dataset and tested with FireNet-Test.
    The KutralNet model is 5.7\% lower than ResNet50 with FireNet, and it gets the best accuracy trained over the FiSmo dataset.
    The ResNet50 version gets better performance with the FiSmoA, followed by KutralNet.
    ]{\includegraphics[width=0.8\columnwidth]{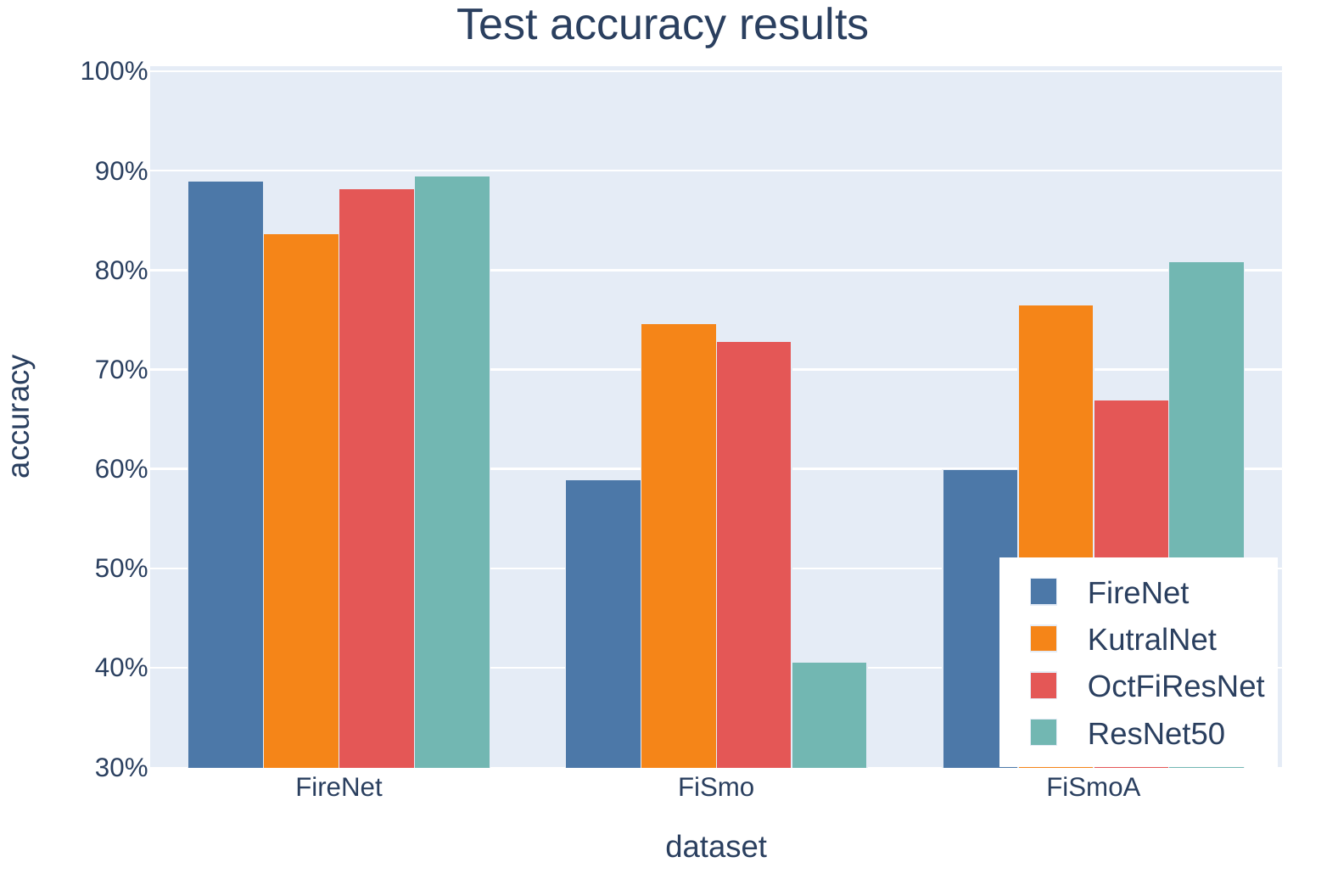} 
    \label{fig:TestResults}
    }
        
    \caption{Training results of datasets FireNet, FiSmo, including FiSmoA, the augmented version of FiSmo (with 485 additional black images). The datasets were split as 70/30 for FireNet and 80/20 for both variants of FiSmo.
    The augmentation with black images allows a better generalization in all the models, but just a slight difference for KutralNet.}
\end{figure*}

% Baseline ROC curves
\begin{figure*}%[htp]
  \centering
  \subfigure[ROC curve for the models trained over the FireNet dataset.
  The FireNet model performs slightly better than the modified version of ResNet50, OctFiResNet, and KutralNet.
  ]{\includegraphics[width=0.3\textwidth]{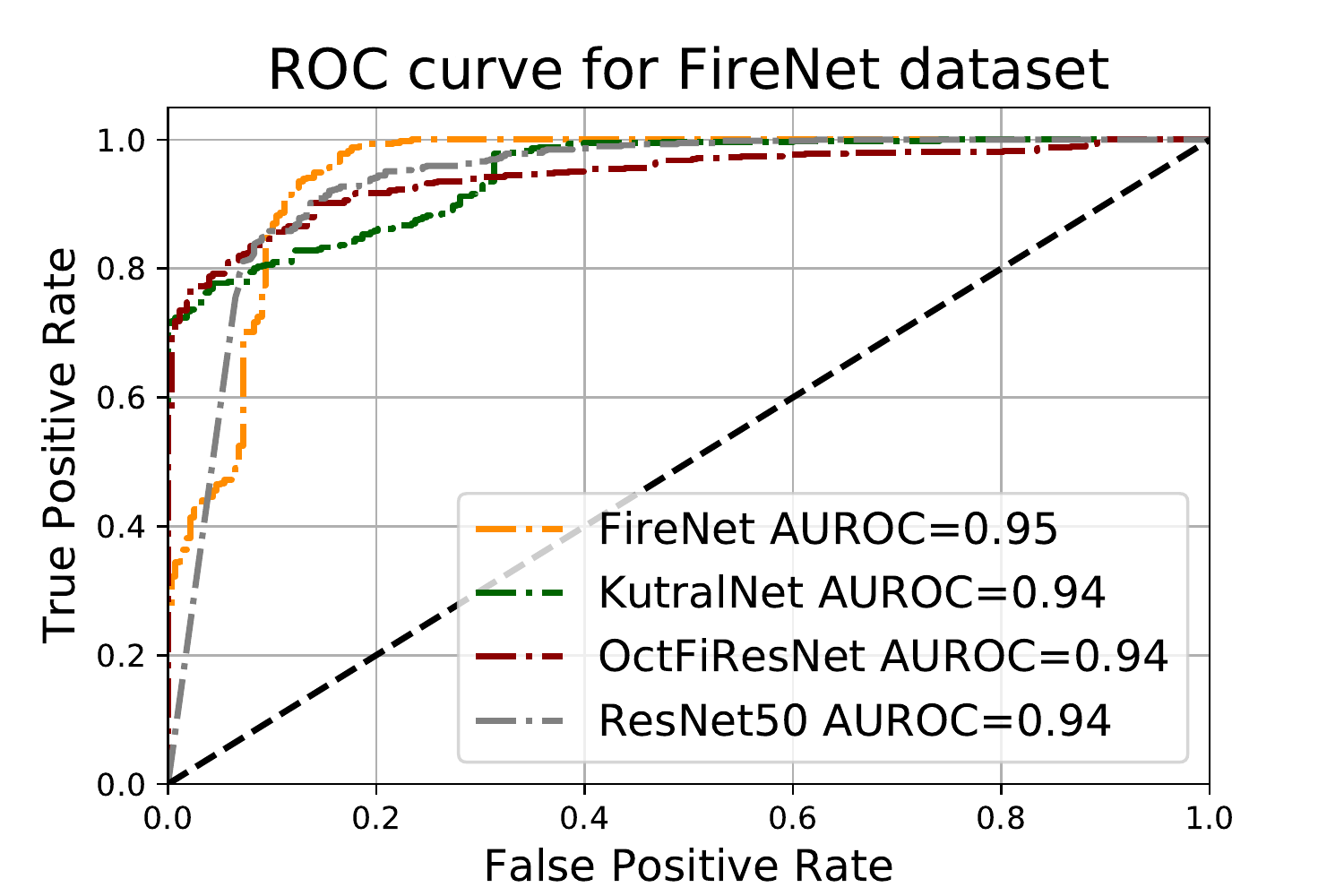} \label{fig:baseline_firenet_test}
  }\quad
  \subfigure[ROC curve for the models trained over the FiSmo dataset.
  FireNet performs the best AUROC value but with low test accuracy.
  The KutralNet model is the second-best AUROC value achieving the best test accuracy, followed by OctFiResNet and ResNet50.
  ]{\includegraphics[width=0.3\textwidth]{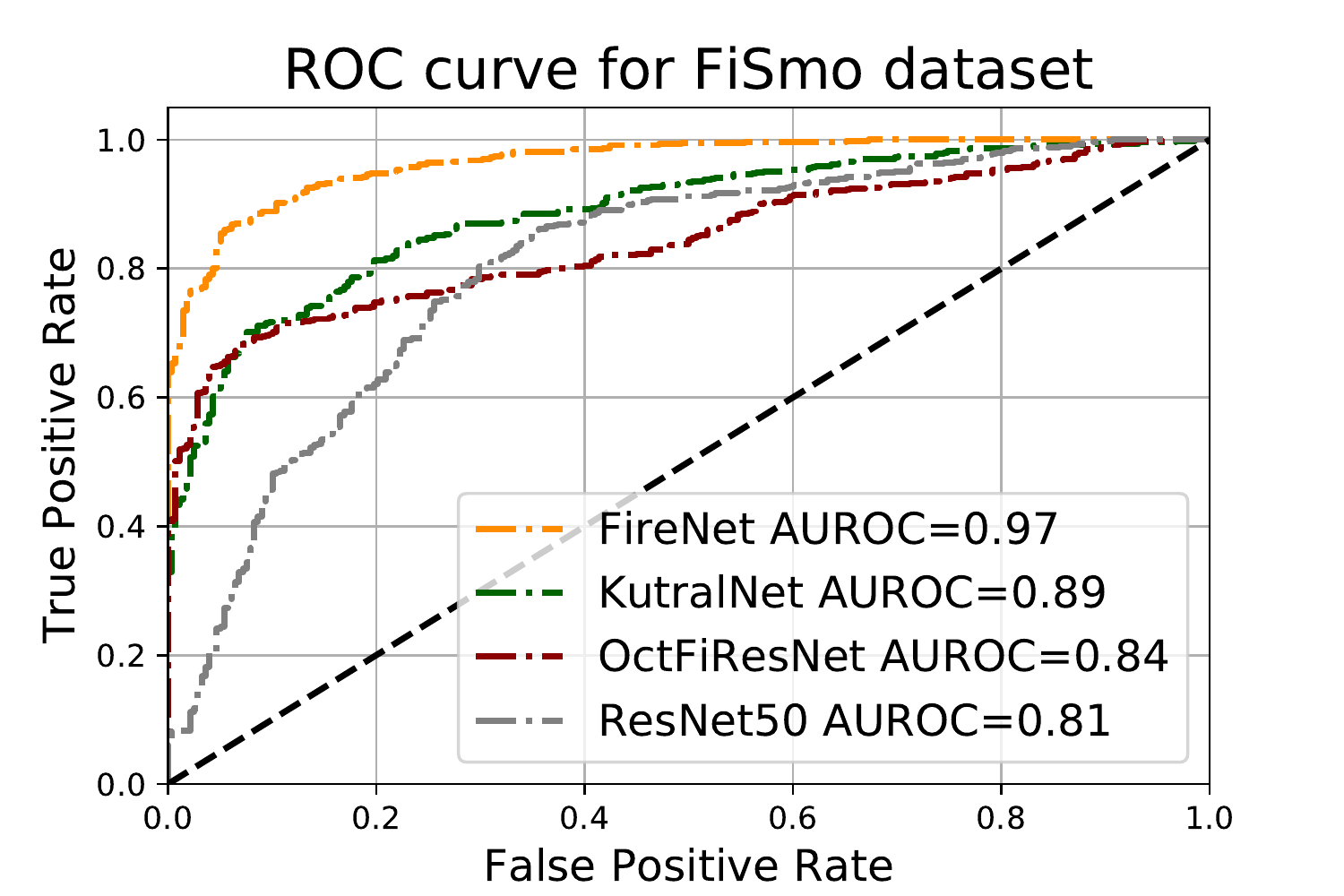} \label{fig:baseline_fismo_test}
  }\quad
  \subfigure[ROC curve for the models trained over the FiSmoA dataset.
  The performance improves in all the models with this augmented version of FiSmo.
  Again, FireNet presents the best AUROC value but with low test accuracy.
  The ResNet50 performs the second-best AUROC value achieving the best test accuracy, followed by KutralNet and OctFiResNet.
  ]{\includegraphics[width=0.3\textwidth]{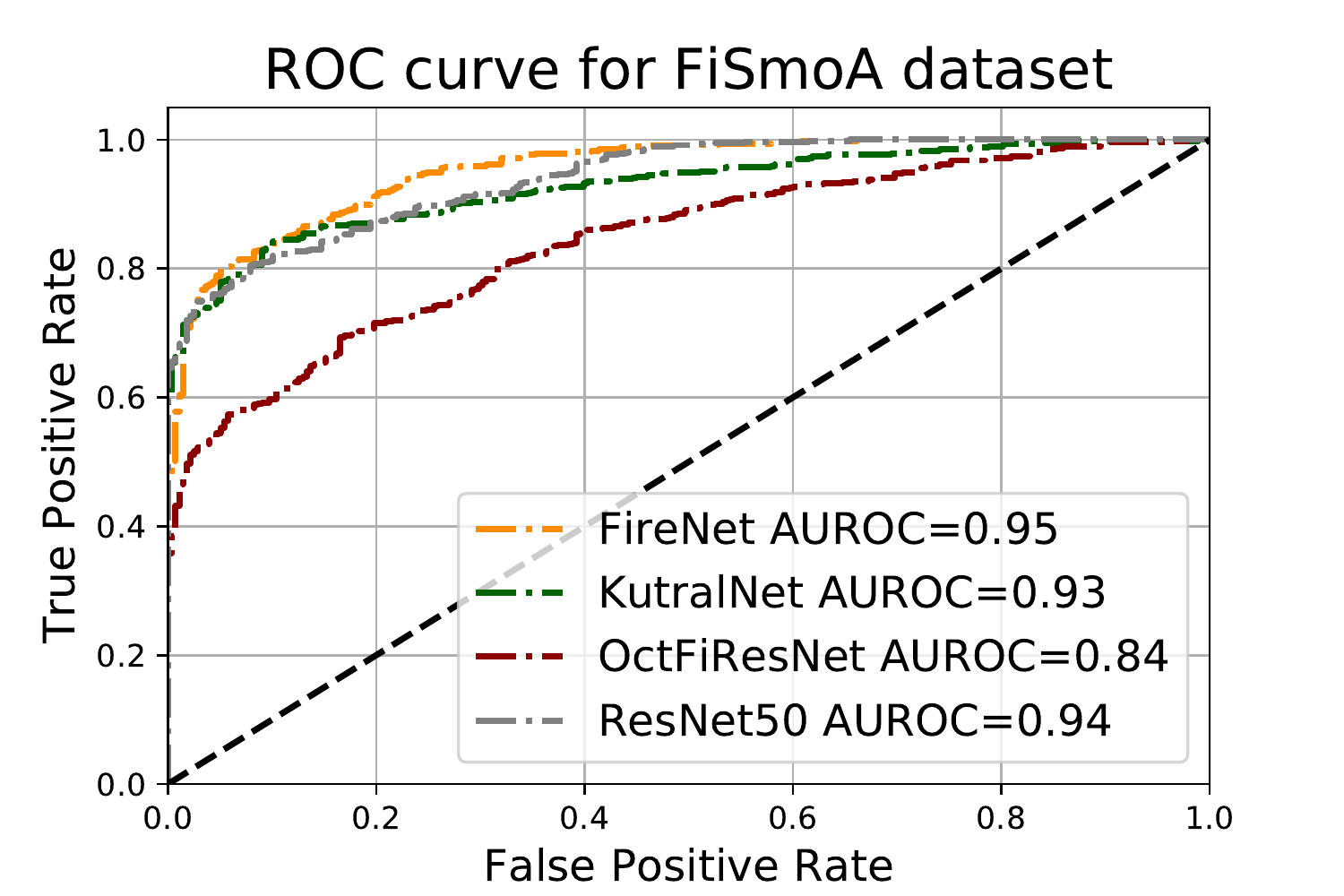} \label{fig:baseline_fismo_black_test}
  }
  
  \caption{The test results of the models with the FireNet-Test dataset with 871 images for fire classification.
  All the models were trained with a different dataset and tested.
  The augmented dataset, FiSmoA, presents better results than FiSmo for all the models.}
  \label{fig:TestMetrics}
\end{figure*}

%%%%%%%%%%%%%%%%%%%%%%%%%%%%%%%%%%%%%%%%%%%%%%%%%%%%%%%%%%%%%%%%%%%%%%%%%%%%%%%%%%%%%%%%%%%%%%%%%%%%%%%%%%%%%%%%%%%%%%%%%%%%%%%%%%%%%%%%%%%%%%%%%%%%%%%%%%%%%%
\section{Results and Discussion}

The next two subsections separate the experiments in order to achieve a portable deep learning model for fire recognition.
The first experimentation was with the proposed baseline model KutralNet, which uses novel deep learning techniques for image classification.
Experimentation results compare KutralNet with other previously presented deep learning models.
After the comparison with our baseline, the final experimentation allowed us to optimize the computational cost of the model, exploring the benefits of different portable approaches presented in the last years as the inverted residual block, the depth-wise, and octave convolution.
The different proposals got almost the same accuracy as the baseline model.

% Portable training Results
\begin{figure*}%[htp]
    \centering
    \subfigure[Validation accuracy obtained from each dataset by different portable models.
    All the models presents almost the same results. ]{\includegraphics[width=0.8\columnwidth]{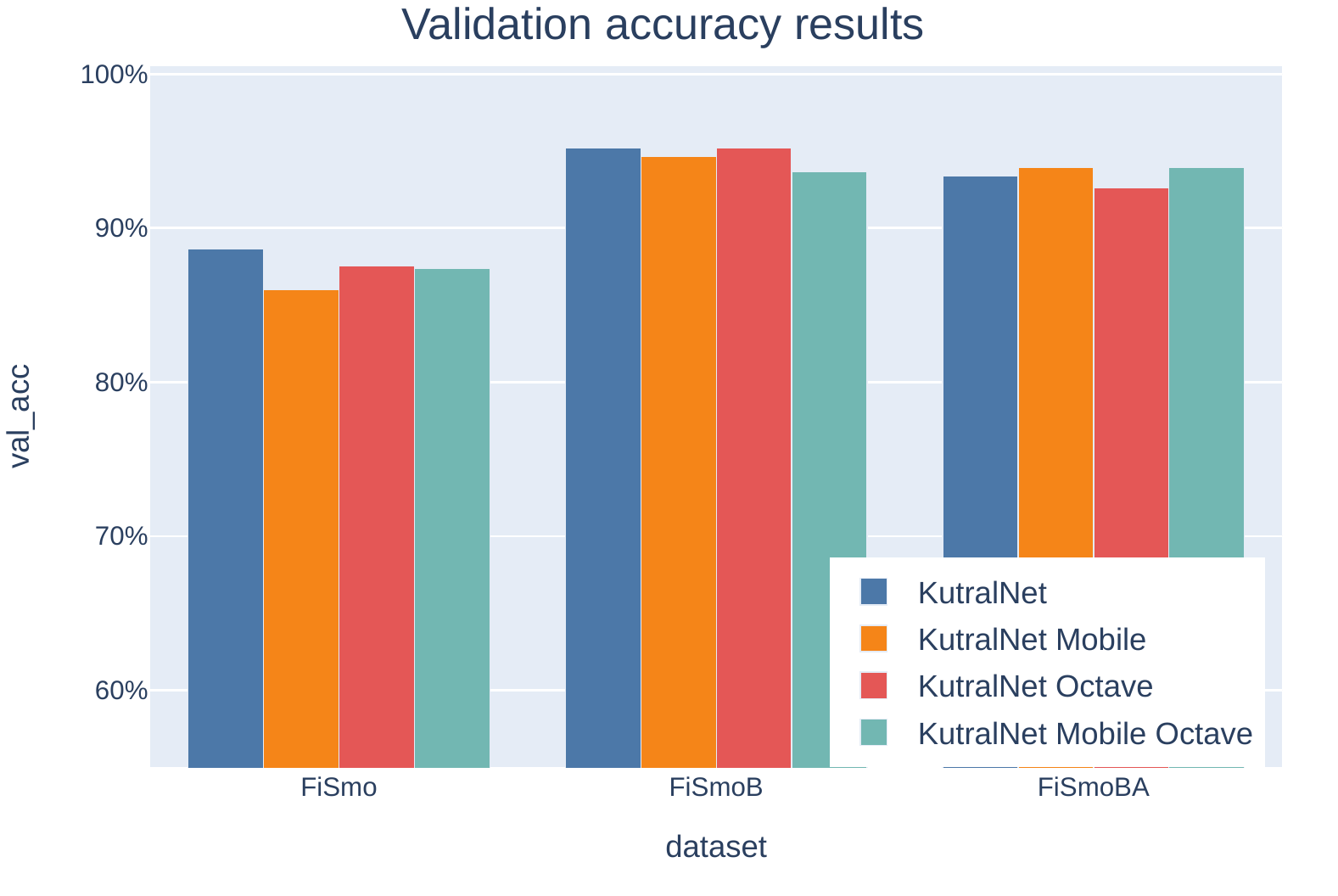}
        \label{fig:PortableTrainingResults}
    }\quad
    \subfigure[Test accuracy of each portable model trained with a different dataset and tested with the FireNet-Test dataset.
    The KutralNet Mobile Octave and KutralNet Octave were capable of outperforming the baseline results.
    ]{\includegraphics[width=0.8\columnwidth]{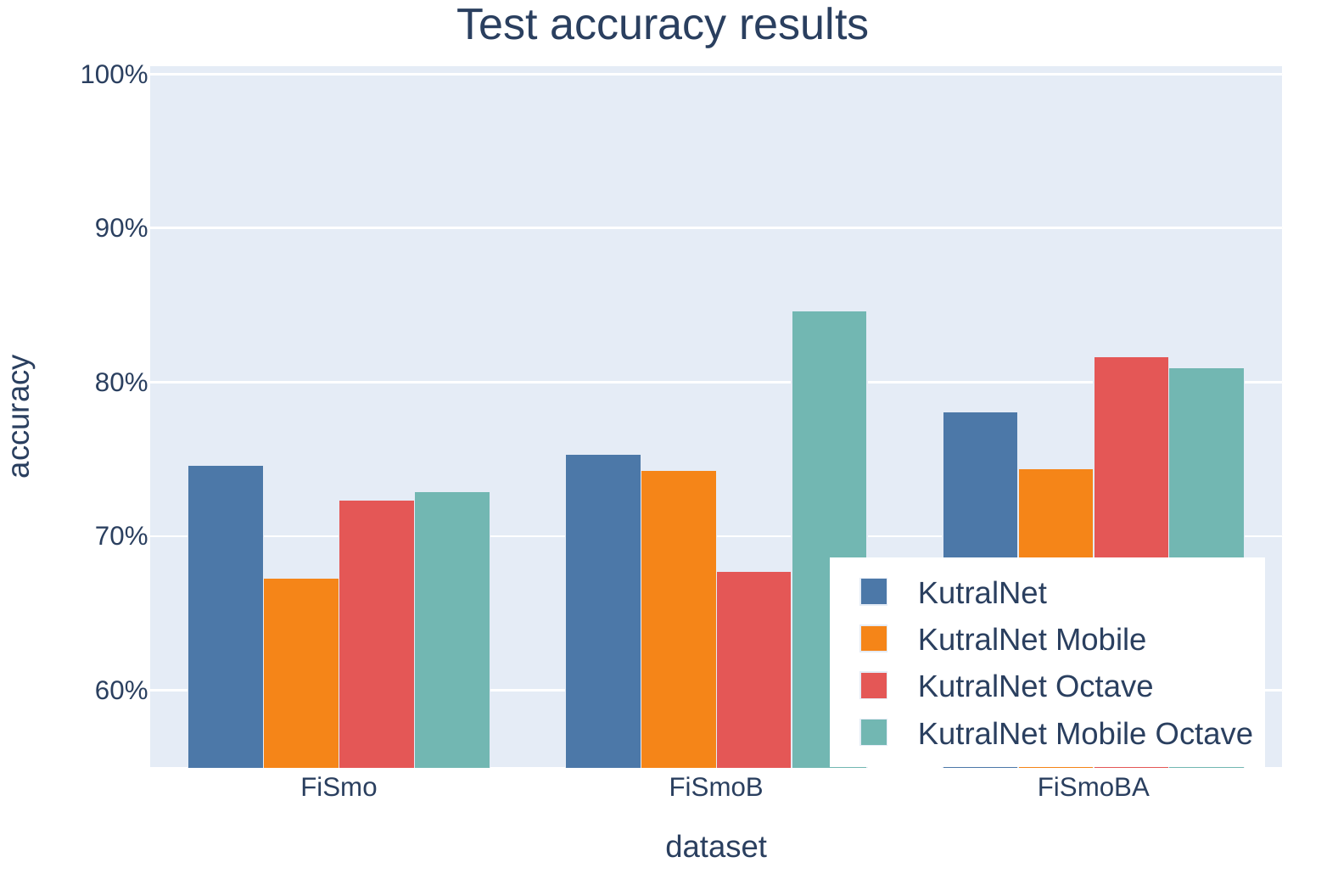} \label{fig:PortableTestResults}}
        
    \caption{Training results obtained with the used datasets: FiSmo, FiSmoB, and FiSmoBA, a variant with 98 replaced no-fire images with black images.
    The datasets were a validation split of 80/20 for all the models.
    The black image augmentation reduces the difference distance between models' accuracy.}
    \label{fig:PortableResults}
\end{figure*}

% Portable ROC curves
\begin{figure*}%[htp]
  \centering
  \subfigure[ROC curve for the models trained over the FiSmo dataset.
  The KutralNet Octave performs the best, followed by KutralNet.
  KutralNet Mobile Octave and the KutralNet mobile models get under the KutralNet's performance.
  ]{\includegraphics[width=0.3\textwidth]{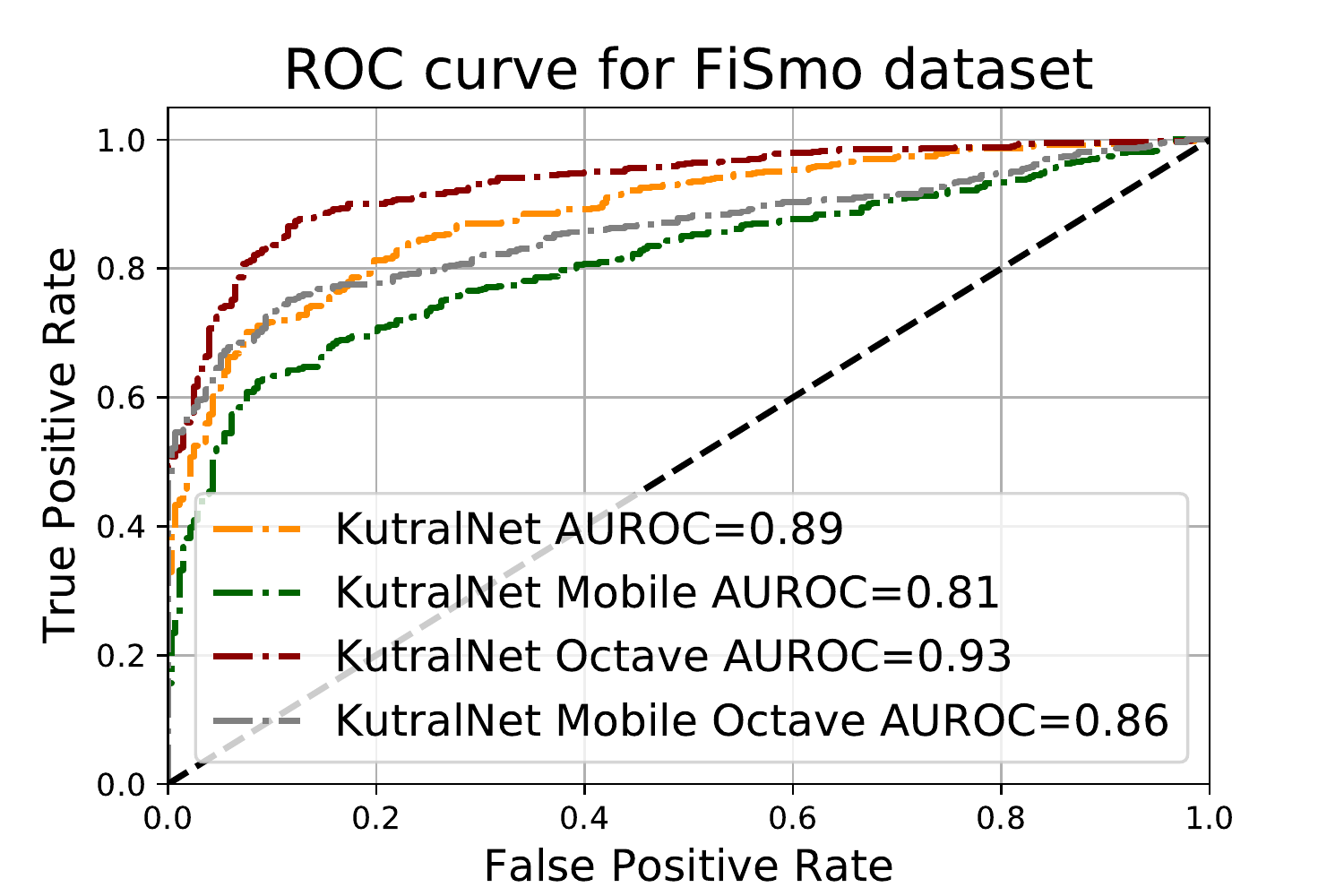} \label{fig:portable_fismo_test}
  }\quad
  \subfigure[ROC curve for the models trained over FiSmo balanced dataset.
  The KutralNet Mobile Octave outperforms the KutralNet results, followed by the Mobile version.
  ]{\includegraphics[width=0.3\textwidth]{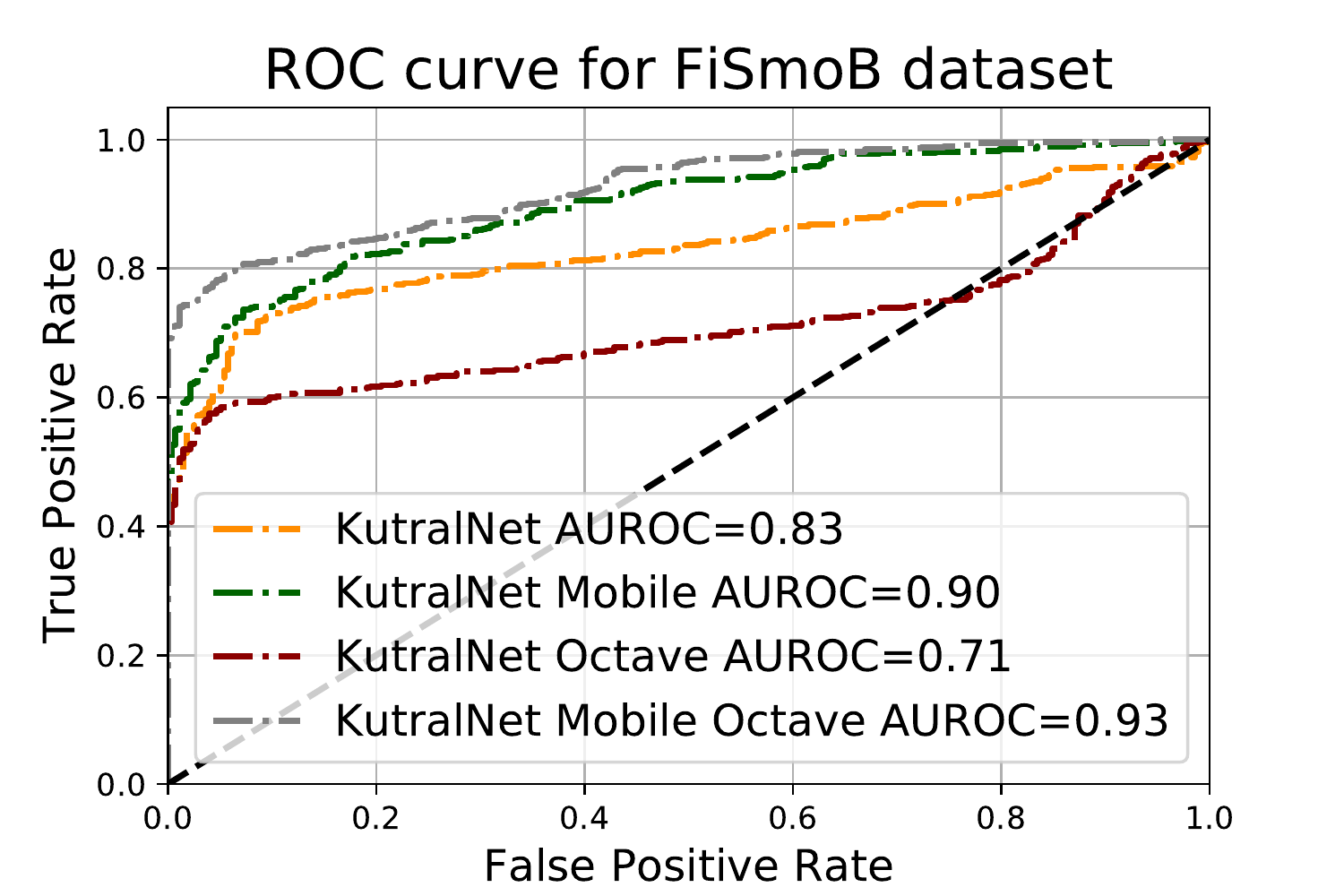} \label{fig:portable_fismo_balanced_test}
  }\quad
  \subfigure[ROC curve for the models trained over the augment balanced FiSmo dataset.
  Just the Octave and Mobile Octave version of KutralNet outperforms the baseline.
  ]{\includegraphics[width=0.3\textwidth]{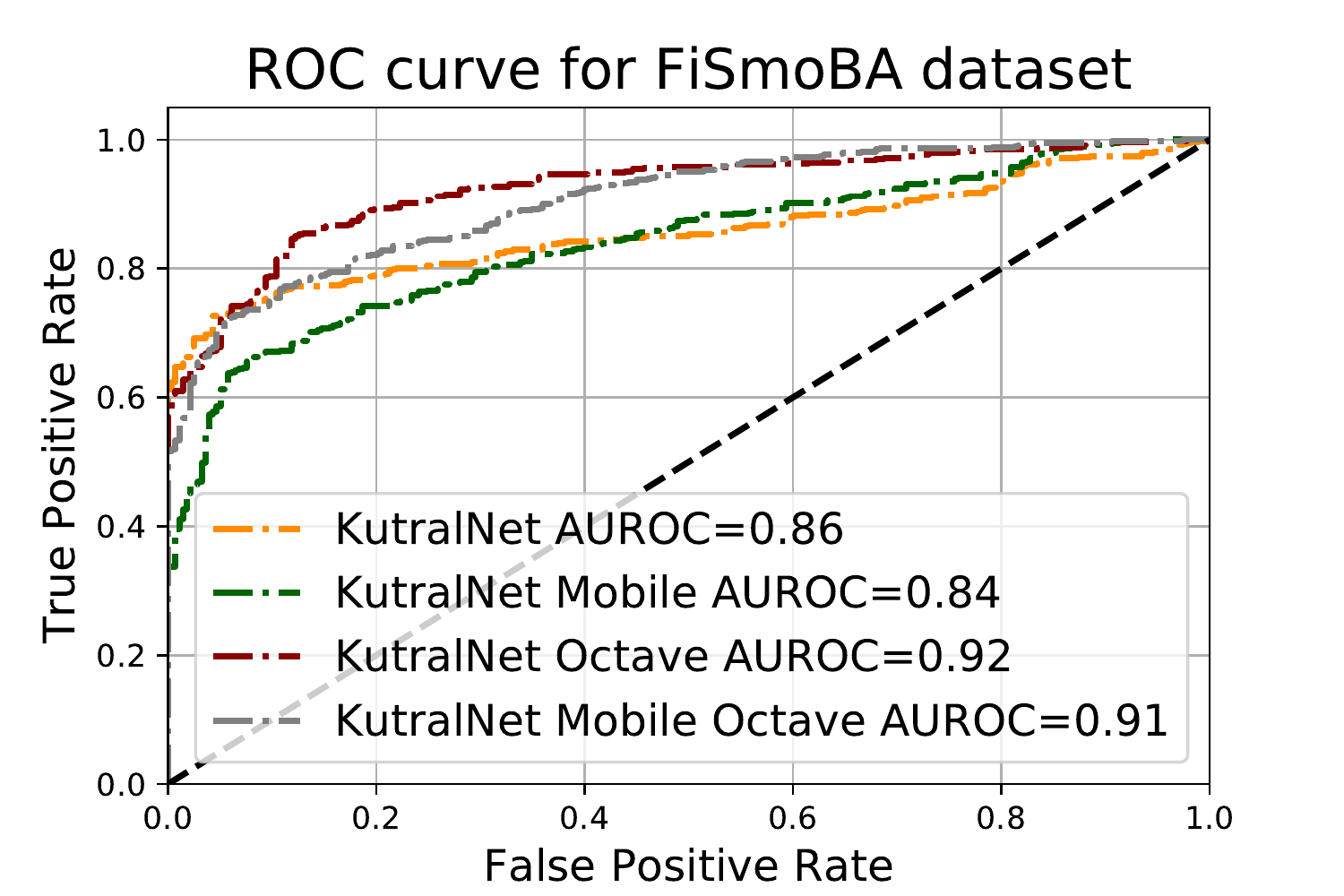} \label{fig:portable_fismo_balanced_black_test}}
  
  \caption{The test results for the portable models with the FireNet-Test dataset with 871 images for fire classification.
The models' results are from the training with different datasets. The FiSmo, FiSmoB, and the FiSmoBA with 98 no-fire images replaced with black images.}
  \label{fig:PortableTestMetrics}
\end{figure*}

%%%%%%%%%%%%%%%%%%%%%%%%%%%%%%%%%%%%%%%%%%%%%%%%%%%%%%%%%%%%%%%%%%%%%%%%%%%%%%
\subsection{Baseline comparison}
The baseline comparison is performed with three deep models, in order to improve the results of our proposed model, focusing on efficiency and lightweight.
The first model is FireNet from Jadon et al. \cite{Jadon2019:FireNet}, which comprises just a few convolution layers and is part of a fire alarm system.
The second model is presented by Sharma et al. \cite{Sharma2017:deepfire}, where a pre-trained ResNet50 makes the feature extraction for a multilayer perceptron classifier with 4096 hidden units.
Finally, the OctFiResNet model \cite{Ayala2019:OctFiResNet} is a reduced version of ResNet50 presenting fewer layers and replacing almost all the vanilla convolution by octave convolution.

The first comparison has been performed over the FireNet dataset with 2425 images, 1124 with fire label, and 1301 with the no-fire label.
The FireNet dataset also contains a test subset with 871 images corresponding to 593 with the fire label and 278 with the no-fire label.
The first results show a validation accuracy of 93.83\%, 96.02\%, 95.34\%, and 98.22\% for FireNet, KutralNet, OctFiResNet, and ResNet50 respectively for the FireNet dataset.
Correspondingly, the test accuracy results are 88.98\%, 83.70\%, 88.18\%, and 89.44\% for FireNet, KutralNet, OctFiResNet, and ResNet50.
In order to check the generalization of the models, for the training and validation, the FiSmo dataset was used.
The test results obtained against the FireNet-Test dataset, as expected, get a lower accuracy for training with the FiSmo dataset.

The following experiment evaluated the prediction of the model with a black image as input.
All the models trained with the FireNet dataset miss-classify the black image, and the same behavior occurs with those models trained with FiSmo, also miss-classifying the black image with some exceptions.
The FiSmo dataset was augmented, to deal with this miss-classification issue, adding a 10\% of the no-fire label images, of black images.
The augmentation for this task showed useful improvements in training and test stage; both results can be observed in Figure \ref{fig:TrainingResults} and Figure \ref{fig:TestResults}.
The most in-depth models outperform the results of the FireNet model.

The test performance of the models trained with different datasets are in Figure \ref{fig:TestMetrics}.
As can be seen, comparing Figure \ref{fig:baseline_fismo_test} to \ref{fig:baseline_fismo_black_test}, the black images added into the FiSmo dataset show a reaction in the behaviour of the ROC curve of the models.
For the KutralNet and ResNet50 models present an improvement, for FireNet a diminishment, and  OctFiResNet remains almost the same.
Regarding to the AUROC index, FireNet achieves better value in all the datasets, but as presented in Figure \ref{fig:TestResults} for FiSmo and FiSmoA achieves a low test accuracy.
To visualize the comparison of the models in a more straightforward way, Table \ref{table:BaselineMeanPerf} shows the average value of test accuracy and, AUROC index for all the datasets of each model.
Our proposed approach presents good accuracy for validation and testing using different datasets.
Overall, KutralNet presents the same behavior as a deep learning model, achieving high performance with a reduced number of parameters and operations.

Our baseline proposed as KutralNet accomplishes an interesting performance compared with previous deep models for fire recognition.
This model presents a few convolution layers in order to acquire a feature representation of fire in images.
A model with a few numbers of layers consequently present a reduced number of parameters and operations required for this task.
Our resultant baseline reduces 85\% the parameters number and 92\% the number of operations required, in comparison to the OctFiResNet model, to process an image signal of 84x84 pixels in RGB channels.

%Tabla con valores medios de performance
\begin{table}%[h]
    \caption{Mean performance values for testing accuracy and AUROC index of each model.}
    \centering
    \begin{tabularx}{\columnwidth}{lYY}
        \toprule
        \textbf{Model} & \textbf{Test Acc} & \textbf{AUROC}\\ 
        \midrule
        \textbf{FireNet} & 64.27\% & \textbf{0.96} \\ 
        \textbf{KutralNet} & \textbf{78.26\%} & 0.92  \\ 
        \textbf{OctFiResNet} & 75.92\% & 0.87 \\ 
        \textbf{ResNet50} & 70.26\% & 0.90 \\ 
        \bottomrule
    \end{tabularx}
    \label{table:BaselineMeanPerf}    
\end{table}

%%%%%%%%%%%%%%%%%%%%%%%%%%%%%%%%%%%%%%%%%%%%%%%%%%%%%%%%%%%%%%%%%%%%%%%%%%%
\subsection{Portable version}
With our KutralNet baseline architecture, the next experimentation was to reduce its computational cost.
For this purpose, from the baseline, some convolution layers are modified, resulting in three different models to check the most efficient way of convolution.
The first model, KutralNet Mobile, replaces the structure of the baseline superficially in order to get the inverted residual blocks with depth-wise convolution, as proposed in \cite{Sandler2018:MobileNet}, simplifying the operations required for the processing.
The second model, KutralNet Octave, replaces the vanilla convolution from the baseline with the octave convolution \cite{Chen2019:octave} for signal processing.
In order for the octave convolution works, the shortcut of the baseline's architecture is slightly modified.
For the third model, KutralNet Mobile Octave is the version that presents a combination of the previously mentioned convolutions.
This version implements the inverted residual block with the octave convolution.

%Tabla con valores medios de performance
\begin{table}%[h]
    \caption{Mean performance values for testing accuracy and AUROC index of each portable model.}
    \centering
    \begin{tabularx}{\columnwidth}{lYY}
        \toprule
        \textbf{Model} & \textbf{Test Acc} & \textbf{AUROC}\\ 
        \midrule
        \textbf{KutralNet} & 76.01\% & 0.86 \\ 
        \textbf{KutralNet Mobile} & 71.99\% & 0.85  \\ 
        \textbf{KutralNet Octave} & 73.90\% & 0.85 \\ 
        \textbf{KutralNet Mobile Octave} & \textbf{79.49\%} & \textbf{0.90} \\ 
        \bottomrule
    \end{tabularx}
    \label{table:PortableMeanPerf}    
\end{table}

In the first place, the training was performed over FiSmo, an unbalanced dataset with 2004 and 4059 images for the fire and no-fire label, respectively.
For this comparison of portable approaches, a validation accuracy of 88.62\%, 85.99\%, 87.55\%, and 87.39\% is achieved by KutralNet, KutralNet Mobile, KutralNet Octave, and KutralNet Mobile Octave model respectively.
Additionally, the test accuracy obtained from the models trained with this dataset is 74.63\%, 67.28\%, 72.33\% and, 72.91\%, respectively.
The bar plot in Figure \ref{fig:PortableResults} shows the results obtained with the other datasets.
For the trained models, the black image test was carried out in order to check the quality of features obtained from the signal.
For this purpose, using the FiSmoBA dataset gets the lowest miss-classification error in all the trained models.
Additionally, it gets $\pm 1\%$ of validation accuracy difference compared with the FiSmoB dataset.
For the black image test, the KutralNet Mobile with octave convolution gets the lowest miss-classification with 10\%, 30\%, and 0\% for the FiSmo, FiSmoB, and FiSmoBA respectively.
In overall, as can be seen in Figure \ref{fig:PortableTestMetrics}, the KutralNet Mobile Octave performs well in the different variations of the FiSmo dataset.
Additionally, the AUROC index is even better than the baseline with the balanced version of the dataset, and the augmented balanced version.
For the case of the KutralNet Octave, it performs better to the Mobile Octave version with the FiSmo and its augmented balanced version.
In Table \ref{table:PortableMeanPerf} are the mean values obtained for test accuracy and AUROC index for all the datasets of each portable model.
Taking into consideration the trade-off between parameter numbers and operations required for processing the image, the Kutralnet Octave presents a suitable solution with a less number of parameters than the KutralNet Mobile Octave and, conversely, requires more operations for processing.

%%%%%%%%%%%%%%%%%%%%%%%%%%%%%%%%%%%%%%%%%%%%%%%%%%%%%%%%%%%%%%%%%%%%%%%%%%%%%%%%%%%%%
\section{Conclusions}
In this work, we have proposed a lightweight approach for fire recognition, which consists of 138.9K parameters and 76.9M flops used as a baseline to build three portable versions.
The KutralNet Mobile, KutralNet Octave, and KutralNet Mobile Octave compare the efficiency of the inverted residual block, the depth-wise convolution, and octave convolution techniques for portable models.

Our proposed model KutralNet obtains better accuracy and AUROC index than previously deep learning models for fire recognition, with just a few layers and with a considerable reduction in computational cost.
The portable version KutralNet Mobile Octave can achieve good performance even if trained with different datasets for the fire and no-fire classification task, requiring only of 24.6M flops with 185.3K parameters.
The computational cost reduction has been possible using the inverted residual block with depth-wise and octave convolution combined for signal processing, proving to be the best approach for feature extraction for portable models.
Additionally, the augmentation with black images improves the generalization in the fire recognition task for deep models, for both balanced and unbalanced datasets.

As future work, we consider applying portable deep learning models for fire recognition and detection on video signal sources.
Furthermore, we plan to extend our work to fire detection using a bounding box approach as well.

%%%%%%%%%%%%%%%%%%%%%%%%%%%%%%%%%%%%%%%%%%%%%%%%%%%%%%%%%%%%%%%%%%%%%%%%%%%%%%%%%%%%%%
\section*{Acknowledgment}
This study was financed in part by the Coordenação de Aperfeiçoamento de Pessoal de Nível Superior - Brasil (CAPES) - Finance Code 001, Fundação de Amparo a Ciência e Tecnologia do Estado de Pernambuco (FACEPE), and Conselho Nacional de Desenvolvimento Científico e Tecnológico (CNPq) - Brazilian research agencies.%, and Universidad Central de Chile under the research project CIP2018009.

\bibliographystyle{./bibliography/IEEEtran}
\bibliography{references}
\balance

\end{document}